\documentclass[sigconf]{acmart}
\usepackage{graphicx}
\usepackage{multirow}
\usepackage{array}
\usepackage{booktabs}
\usepackage{caption}
\usepackage{enumitem}
\usepackage{hanging}
\usepackage{hyperref}
\usepackage{algorithm}
\usepackage{algpseudocode}
\usepackage{amsmath}
\usepackage{booktabs}
\usepackage{subcaption}
\usepackage{array} % 增强型表格功能
\usepackage{tabularx}
\definecolor{lightblue}{rgb}{0.21, 0.49, 0.74}

\usepackage{makecell} % For \makecell
\usepackage{tcolorbox} % 引入 tcolorbox 包
\usepackage{lipsum} % 引入 lipsum 包，用于创建示例图像
\usepackage{tikz} % 引入 TikZ 包，用于创建示例图像
\AtBeginDocument{%
  }
\copyrightyear{2025}
\acmYear{2025}
\setcopyright{acmlicensed}
\acmConference[MM '25]{Proceedings of the 33rd ACM International Conference on Multimedia}{October 27--31, 2025}{Dublin, Ireland}
\acmBooktitle{Proceedings of the 33rd ACM International Conference on Multimedia (MM '25), October 27--31, 2025, Dublin, Ireland}
\acmDOI{10.1145/3746027.3755011}
\acmISBN{979-8-4007-2035-2/2025/10}
\begin{document}

%%
%% The "title" command has an optional parameter,
%% allowing the author to define a "short title" to be used in page headers.
\title{SVGen: Interpretable Vector Graphics Generation with Large Language Models}

%%
%% The "author" command and its associated commands are used to define
%% the authors and their affiliations.
%% Of note is the shared affiliation of the first two authors, and the
%% "authornote" and "authornotemark" commands
%% used to denote shared contribution to the research.
\settopmatter{authorsperrow=4}
\author{Feiyu Wang}
\authornote{Work done during an internship at TeleAI}
\authornote{Equal Contribution}
\affiliation{%
  \institution{Northwestern Polytechnical
University}
  \city{Xi’an}
  \country{China}
}
\affiliation{%
  \institution{Institute of Artificial Intelligence (TeleAI), China Telecom}
  \city{Beijing}
  \country{China}
}
\email{wfy@mail.nwpu.edu.cn}

\author{Zhiyuan Zhao}
\authornotemark[2]
\affiliation{%
  \institution{Institute of Artificial Intelligence (TeleAI), China Telecom}
  \city{Beijing}
  \country{China}
}
\email{tuzixini@gmail.com }

\author{Yuandong Liu}
\affiliation{%
  \institution{Northwestern Polytechnical
University}
  \city{Xi’an}
  \country{China}
}
\email{LYD2022@mail.nwpu.edu.cn}

\author{Da Zhang}
\affiliation{%
  \institution{Northwestern Polytechnical
University}
  \city{Xi’an}
  \country{China}
}
\affiliation{%
  \institution{Institute of Artificial Intelligence (TeleAI), China Telecom}
  \city{Beijing}
  \country{China}
}
\email{dazhang@mail.nwpu.edu.cn}

\author{Junyu Gao}
\authornote{Corresponding Author}
\affiliation{%
  \institution{Northwestern Polytechnical
University}
  \city{Xi’an}
  \country{China}
}
\affiliation{%
  \institution{Institute of Artificial Intelligence (TeleAI), China Telecom}
  \city{Beijing}
  \country{China}
}
\email{gjy3035@gmail.com}

\author{Hao Sun}
\affiliation{%
  \institution{Institute of Artificial Intelligence (TeleAI), China Telecom}
  \city{Beijing}
  \country{China}
}
\email{sunh10@chinatelecom.cn}

\author{Xuelong Li}
\authornotemark[3]
\affiliation{%
  \institution{Institute of Artificial Intelligence (TeleAI), China Telecom}
  \city{Beijing}
  \country{China}
}
\email{xuelong_li@ieee.org}

%%
%% By default, the full list of authors will be used in the page
%% headers. Often, this list is too long, and will overlap
%% other information printed in the page headers. This command allows
%% the author to define a more concise list
%% of authors' names for this purpose.
\renewcommand{\shortauthors}{Wang et al.}

%%
%% The abstract is a short summary of the work to be presented in the
%% article.
\begin{abstract}
Scalable Vector Graphics (SVG) has become an indispensable technology in front-end development and UI/UX design, due to its inherent advantages in scalability, editability, and rendering efficiency. In the creation of vector graphics, while expressing creative concepts is straightforward, translating them into precise digital artworks is often challenging and time-consuming. To overcome this technical bottleneck and achieve intelligent conversion from concept to final product, we have constructed SVG-1M, a large-scale dataset of high-quality SVG samples with paired textual descriptions. Through innovative data augmentation and annotation processes, we built precisely aligned "Text instruction-SVG code" training pairs, with a subset enhanced by Chain-of-Thought (CoT) annotations. This provides rich semantic supervision signals for model learning. Based on this dataset, we propose SVGen, an end-to-end generative model capable of directly converting natural language descriptions into SVG code. This design addresses the challenges of generating semantically accurate vector graphics while preserving complete structural information. We explored various training strategies and introduced a progressive curriculum learning approach, optimized with reinforcement learning algorithms. Notably, this study innovatively applies the CoT paradigm to vector graphics generation, effectively enhancing both the accuracy and interpretability of SVG synthesis. Experimental validation demonstrates that SVGen exhibits significant advantages over general large models in terms of SVG generation quality, while also surpassing optimization-based rendering methods in generation efficiency. The proposed method enables intelligent conversion between natural language and vector graphics, enabling novel workflows like real-time AI-assisted design iteration. Code, model, and data is released at: https://github.com/gitcat-404/SVGen
\end{abstract}

%SVGen's working process: The model processes user text input through multi-step Chain-of-Thought (CoT) reasoning to parse semantic intent and construct design logic. It progressively deduces graphical elements (shapes, layout, style, and spatial relationships) to generate high-quality, well-structured SVG icons that adhere to design specifications.
\begin{teaserfigure}
  \includegraphics[width=\textwidth]{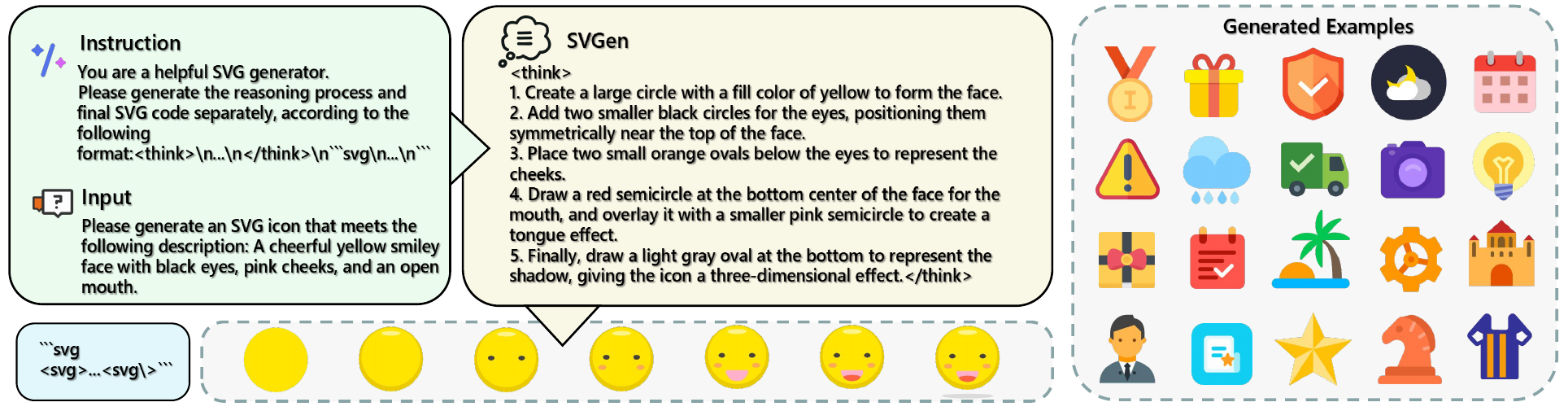}
  \captionsetup{justification=raggedright} % 设置 caption 左对齐
  \caption{SVGen analyzes the creative requirements of users, gradually deriving the design logic to generate high-quality SVG icons that meet design standards.}
  \label{fig:teaser}
\end{teaserfigure}

%%
%% The code below is generated by the tool at http://dl.acm.org/ccs.cfm.
%% Please copy and paste the code instead of the example below.
%%


%%
%% Keywords. The author(s) should pick words that accurately describe
%% the work being presented. Separate the keywords with commas.
%% A "teaser" image appears between the author and affiliation
%% information and the body of the document, and typically spans the
%% page.

%%
%% This command processes the author and affiliation and title
%% information and builds the first part of the formatted document.
\maketitle

\section{Introduction}
With the rapid advancement in front-end development and user interface/user experience design, Scalable Vector Graphics (SVG)\cite{quint2003scalable,peng2000scalable} has become an essential tool due to its unique advantages in scalability, editability, and efficient rendering.\cite{ferraiolo2000scalable} However, translating abstract design intentions into precise SVG digital works still faces dual challenges: On the one hand, traditional design tools rely on professionals performing complex manual operations to encode graphics; on the other hand, existing AI generation methods are often limited to bitmap outputs, making it difficult to preserve the structured features and editing attributes inherent to vector graphics. This bottleneck from concept to code significantly constrains design efficiency, particularly in scenarios requiring rapid iteration.\cite{han2015two} Simplifying SVG generation to better enable designers to express their creativity has thus become a research hotspot.

Current research on SVG generation primarily focuses on two major directions: image vectorization and Text-to-SVG generation. Image vectorization aims to convert pixel-based raster images into SVG representations, which is a fundamental challenge in the field of vector graphics\cite{belouadi2023automatikz,rodriguez2023figgen,rodriguez2023ocr}. Traditional image processing methods typically approximate images through multiple paths\cite{visioncortex2023,selinger2024,weber2024}, while deep learning methods utilize latent variable models and differentiable rendering techniques for advanced vector graphic modeling \cite{cao2023svgformer,wang2021deepvecfont}. Recent advances\cite{rodriguez2023starvector,zhang2023beyond} have introduced Multi-modal Large Language Models (MLLMs), treating image vectorization as an inverse rendering and code generation task, thereby promoting flexible conversion between bitmaps and vector graphics and significantly enhancing the richness and applicability of SVGs.

In recent years, research on Text-to-SVG generation has gained momentum, presenting two primary technical approaches: path sequence-based methods and optimization-based methods. 
In the optimization-based approach, recent studies\cite{thamizharasan2024nivel,xing2023diffsketcher,zhang2024text} employ diffusion model\cite{rombach2022high} to generate raster images, which are then converted into SVG representations using differentiable rasterizers\cite{li2020differentiable}. SVGs obtained in this way are generally not editable and difficult to maintain.
% Although diffusion models excel in natural image generation, they face notable limitations when applied to SVG icon generation, such as imprecise geometric shapes, blurry distortions, poor editability of output results, and inefficient iterative generation processes, failing to meet professional design requirements.
For path sequence generation, existing methods treat SVG graphics as sequences of basic path commands and use autoregressive methods for sequence prediction\cite{wu2023iconshop,chen2024svgbuilder}. While these methods maintain vector editability, they are constrained by the semantic understanding capabilities of models and the expressive power of basic path commands\cite{jiang2018deep}, limiting them to generating simple graphics or specific categories of content (such as basic shapes and fonts), which fall short in handling complex scenarios.

% Meanwhile, Large Language Models (LLMs)\cite{radford2019language,touvron2023llama,zhao2023survey} have demonstrated remarkable capabilities in multi-modal understanding\cite{achiam2023gpt,liu2023visual,wang2024qwen2} and code generation tasks\cite{li2023starcoder,roziere2023code,hui2024qwen2}, offering new perspectives for addressing these issues. 
In conclusion, current research faces two key limitations: (1) the lack of high-quality, large-scale, and semantically aligned "Text-SVG" pairs; and (2) existing generation methods struggle to simultaneously ensure the structural integrity and semantic accuracy of vector graphics. These limitations result in common quality issues such as missing paths and layer misalignment in model outputs, failing to meet practical design needs. To address these challenges, we propose an innovative solution by constructing a large-scale, finely annotated "Text instruction-SVG" dataset and fine-tuning advanced models like Llama3.1\cite{grattafiori2024llama}, Qwen2.5\cite{yang2024qwen2}, and Starcoder2\cite{lozhkov2024starcoder}, incorporating Chain-of-thought\cite{wei2022chain} reasoning and reinforcement learning algorithms to enhance model interpretability and structural representation completeness. Ultimately, we developed SVGen, an end-to-end Text-to-SVG generation model. The core contribution includes:

% 这里不应该是 benchmark，然后 dataset 为什么包含 cot 技术？
\textbf{(1) SVG-1M Dataset:} Constructing the large-scale, high-quality "Text instruction-SVG" aligned dataset with multi-level semantic annotations and Chain-of-thought enhancement technologies, setting a new standard for vector graphic understanding.

%\textbf{(2) Innovative Method Integration:} Introducing chain-of-thought reasoning into the field of vector graphic generation for the first time, combined with innovative reinforcement learning strategies, achieving explainable generation from natural language to structured SVG code.
\textbf{(2) Construct the Progressive Learning Strategy: } From simple monochrome to complex multicolored SVGs, combined with three different levels of textual description (concise, detailed, and specific implementation steps). We design "curricula" of varying difficulty for the model to gradually enhance its understanding of SVG graphics. In addition, we design the path number matching reward and the integrity rewards and combine them with reinforcement learning algorithms to further constrain the quality of generated SVG code.

\textbf{(3) End-to-End SVG Generation Framework:} This work develops an SVG generation model based on a lightweight LLM that efficiently generates corresponding SVG code according to user requirements. Relevant experiments also demonstrate that its performance surpasses the optimization-based methods and LLMs with significantly more parameters.
%Developing an LLM-based SVGen model capable of directly producing production-grade SVG code, significantly improving generation efficiency and quality while maintaining the structural integrity of vector graphics, surpassing existing optimization-based methods.

\section{Related Work}

\subsection{Large Language Models}
Large Language Models (LLMs) have made significant strides in the field of natural language processing, giving rise to a series of prominent models such as OpenAI's GPT-4\cite{achiam2023gpt}, Meta's Llama\cite{touvron2023llama}, Anthropic's Claude 3.7\cite{anthropic_claude_sonnet}, Alibaba's Qwen 2.5\cite{yang2024qwen2}, Google's Gemini 2.5 Pro\cite{gemini_20}, xAI's Grok-3\cite{grok3_beta}, and DeepSeek R1\cite{guo2025deepseek} from DeepSeekAI. These models, through extensive pre-training and instruction fine-tuning, exhibit powerful generalization and multitasking capabilities, finding widespread applications in text generation, code writing, and mathematical reasoning, among other areas. As the technology matures and domain-specific needs grow, researchers are exploring ways to adapt these models to highly specialized fields\cite{gao2025combining,gao2020feature}. By integrating domain-specific datasets and knowledge bases\cite{gao2024nwpu,gao2024imbalanced}, LLMs have demonstrated significant advantages in specialized domains such as finance, medicine, and law, leading to the emergence of specialized models like BloombergGPT\cite{wu2023bloomberggpt} in finance, PMC-LLAMA\cite{wu2024pmc} in medicine, and Disc-lawllm\cite{yue2023disc} in the legal field. These domain-specific models enhance accuracy and practicality in professional tasks by incorporating domain knowledge and specific data, showcasing the immense potential of LLMs in vertical fields. In this task, we aim to leverage the deep semantic understanding and code generation capabilities of large language models to establish a foundational knowledge base for text-driven vector graphic generation, thereby achieving higher levels of generation precision.

\subsection{Vector Graphics Generation}

Early research in the field of vector graphic generation primarily focused on using traditional techniques, such as segmentation and polynomial curve fitting, to vectorize images. DiffVG\cite{li2020differentiable}, as the first differentiable vector graphics rasterization pipeline, pioneered the application of deep learning to vector graphic generation. With the rapid advancement of deep learning, researchers have proposed a variety of new methods. For instance, SVG-VAE\cite{radford2019language} leverages class-conditional Variational AutoEncoders (VAE)\cite{kingma2013auto} to generate vector graphics; DeepSVG\cite{carlier2020deepsvg} adopts a hierarchical VAE architecture combined with Transformer models to represent SVG paths; Im2Vec\cite{reddy2021im2vec} uses Recurrent Neural Networks (RNN) to convert pixel images into latent representations and decodes them into paths. Additionally, StarVector\cite{rodriguez2023starvector}, based on Multi-modal Large Language Models (MLLM), can directly generate SVG code from input images.

In the domain of Text-to-SVG generation, IconShop\cite{wu2023iconshop} is a text-guided vector icon synthesis method that employs autoregressive Transformers, focusing on generating simple monochrome SVG icons. Recently, optimization-based approaches have also been introduced to SVG generation. For example, CLIPDraw\cite{frans2022clipdraw} and CLIPasso\cite{vinker2022clipasso} utilize CLIP\cite{radford2021learning} to optimize sketches for SVG generation, while models like VectorFusion\cite{jain2023vectorfusion} and SVGDreamer\cite{xing2024svgdreamer} leverage the powerful bitmap generation capabilities of diffusion models to iteratively produce SVGs. However, SVG icons generated through optimization-based methods often lack editability and their iterative nature limits generation efficiency.

Although these methods perform well in specific scenarios, they generally exhibit limited expressiveness and suffer from overfitting issues when handling complex generation tasks, thereby constraining their generalization ability. To address these challenges, this paper proposes an SVG generation method based on large language models. By directly rendering vector graphics within the SVG code space, this approach overcomes the aforementioned limitations, achieving significant improvements in both generation speed and quality.

\subsection{SVG datasets}
Regarding SVG datasets, several significant collections have been developed to support related research. For instance, the FIGR-8-SVG\cite{clouatre2019figr} contains 1.5 million monochrome vector icons, all transformed into a unified representation with discretized parameters. The SVG-Icons8\cite{carlier2020deepsvg} comprises 100,000 SVG icons, spanning 56 different categories of diverse real-world graphics, designed to support vector graphics generation and representation learning with consistent scale, color, and style. To address the lack of color information in existing datasets, the ColorSVG-100K\cite{chen2024svgbuilder} dataset has emerged, containing 100,000 richly colored SVGs, aimed at filling the gap in colored SVG generation research. The LLM4SVG\cite{xing2024empowering} project has further enriched resources by collecting and standardizing approximately 250,000 colorful and complex vector graphics, utilizing models like BLIP\cite{li2022blip}, GPT-4, etc., to generate corresponding subtitles for SVGs as text prompts, advancing research in Text-to-SVG conversion. Additionally, the SVG-Stack\cite{rodriguez2023starvector} is a large-scale dataset offering over 2 million raw SVG samples, each paired with a corresponding text description. However, it has drawbacks such as inconsistent canvas sizes and redundant metadata. 
Although existing datasets have advanced research to some extent, they still exhibit various limitations and fall short of fully meeting the requirements for high-quality SVG generation. 
%To address this, We present a large-scale, rigorously normalized Text-SVG paired dataset with structured Chain-of-Thought annotations, which facilitates advancements in SVG generation research.

\section{SVG-1M Dataset Construct}
% To address the limitations of existing datasets when using LLMs for SVG generation, such as insufficient data scale, overly simplistic structures, and imprecise semantic descriptions, we construct a large-scale, rigorously normalized Text-SVG paired dataset named SVG-1M. 
% This dataset comprises 1 million high-quality "Text instruction-SVG code" pairs aimed at achieving precise alignment between icon descriptions and actual semantics, provide a rich set of samples for the subsequent development in this field.

To overcome the scale, structural, and semantic limits of existing SVG datasets for LLMs, we introduce SVG-1M, a large, well-normalized dataset of 1 million high-quality Text–SVG pairs designed to ensure precise "Text instruction-SVG code" alignment and support future research.

\begin{figure}[htbp]
    \centering
    \includegraphics[width=0.48\textwidth]{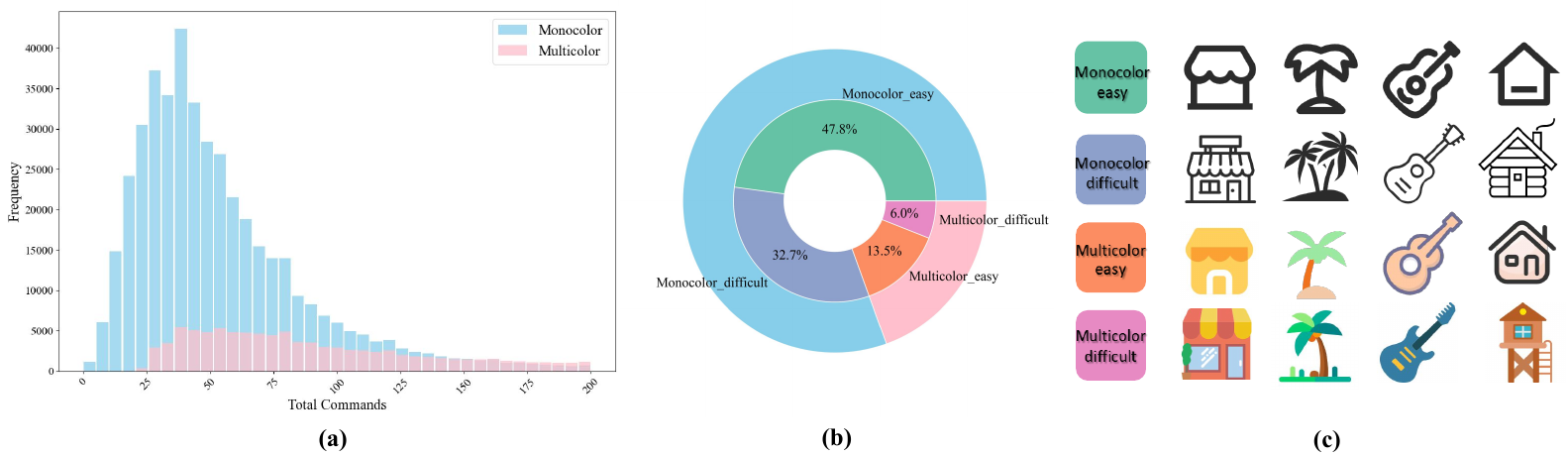}
    \caption{The distribution of data complexity and samples in SVG-1M. Figure (a) shows the command quantity distribution of monochrome and multicolor data. (b) shows the proportion of different complexity levels in the dataset, and (c) presents examples from each level, highlighting differences in geometric design and color styles.}
    \label{fig:2}
\end{figure}

\textbf{Data Collection.} Iconfont\cite{iconfont_cn} is a professional vector icon repository that offers a diverse collection of SVG resources, including emojis, illustrations, fonts, and office icons. All icons on this platform strictly adhere to unified technical specifications: they use standard SVG namespaces, are based on a canvas size of 1024×1024 pixels, and maintain clean path data structures (without redundant metadata). These highly standardized features make it an ideal benchmark dataset for research. We collected approximately 500,000 SVG format icons from Iconfont, including 400,000 monochrome icons and 100,000 colored icons. To ensure data consistency and usability, we performed standardization processes such as converting relative paths to absolute paths and simplifying operation commands into three basic types: "M" (move), "L" (line), and "C" (curve). 
As shown in Table~\ref{tab:command_categories}, we classify SVGs into four difficulty levels based on the number of commands and the types of colors involved. More detailed data distribution can be found in Figure~\ref{fig:2}.
%Based on the number of these commands, we proposed a method to quantify SVG complexity (see Table 1) for subsequent sample classification and analysis. The specific composition of the data is illustrated in Figure 2.

\textbf{Data Inference.} To fully leverage the dual characteristics of SVG icons (structured data and visual representation), we convert selected icons into raster images with a resolution of 200×200 pixels and use GPT-4o to generate detailed image descriptions. This ensures that the appearance of icons can be accurately reproduced solely based on the descriptions. For monochrome icons, we employ small visual language models (such as Qwen2.5-VL-7B\cite{bai2025qwen2}) for additional inference to generate coarse-grained simple descriptions. For colored icons, we utilize data augmentation techniques such as path swapping and color replacement to generate 203,205 SVGs based on the original 100,000 colored icons.
%这块，我觉得应该加一小节文字，描述一下你 cot 数据具体是怎么采集的，要是和 svg 本身紧密相关的思维链设计（要求）
On this basis, we conduct fine-grained inference for all data and apply Chain-of-thought annotations using GPT-4o to 65,745 data pairs, ultimately resulting in 1 million high-quality Text-SVG pairs.

The final SVG-1M dataset comprises three distinct subsets: 826,326 monochrome icon-instruction pairs, 137,460 multicolor icon-instruction pairs, and 65,745 multicolor icon-instruction pairs annotated with CoT reasoning.
It adopts a direct question-answer format where natural language instructions correspond to either raw SVG code or Chain-of-thought + SVG code. This approach preserves complete semantic and structural information, fully utilizes the capabilities of pre-trained models, and significantly simplifies the data processing workflow. Experiments demonstrate that this dataset effectively enhances model performance in visual semantic understanding and SVG code generation tasks, providing a solid foundation for related research.
%For more detailed information on data processing and example samples, please refer to the supplementary materials.

\begin{table}[h!]
\centering
\resizebox{0.95\columnwidth}{!}{ 
\begin{tabular}{l|c|l}
\hline
\textbf{Color Category} & \textbf{Command Count} & \textbf{Difficulty Level} \\ \hline
Monochrome          & 0 - 50                & Monocolor\_easy \\ 
Monochrome          & 50 - 200              & Monocolor\_difficult \\
Multicolor          & 0 - 100               & Multicolor\_easy \\ 
Multicolor          & 100 - 200             & Multicolor\_difficult \\ \hline
\end{tabular}
}
\caption{Classification criteria for SVGs based on color category and the number of commands.}
\label{tab:command_categories}
\end{table}

\begin{figure*}[!thbp]
    \centering
    \includegraphics[width=0.95\textwidth]{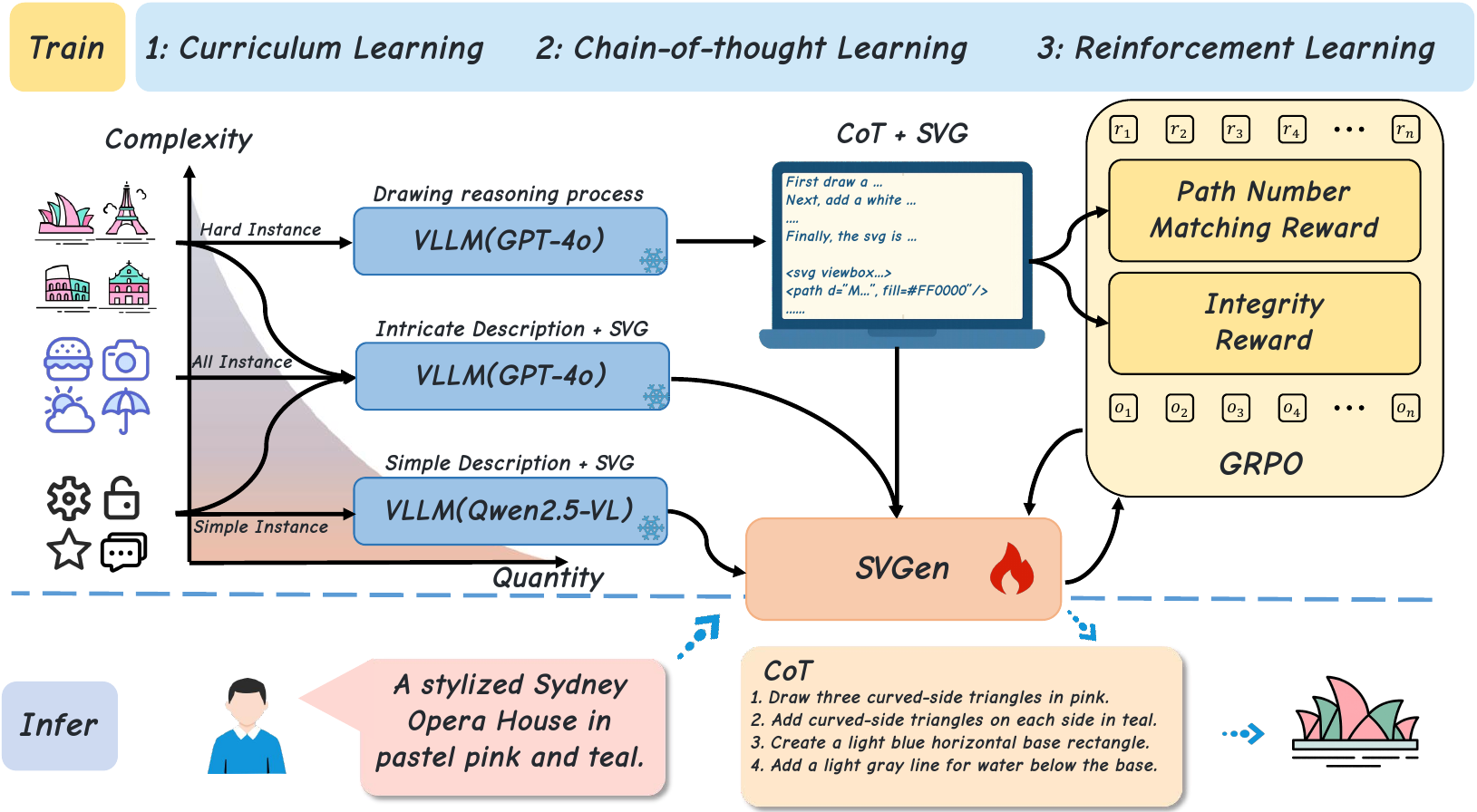}
    \caption{The workflow of SVGen. We start with curriculum learning, where the model gradually trains on SVGs increasing in complexity, from simple monochrome shapes to intricate, colorful graphics. Next, we use complex samples and CoT data to guide the model to provide a step-by-step design process before outputting the SVG. Finally, through reinforcement learning via GRPO, the model is guided to produce structurally complete and visually accurate SVGs by rewarding integrity and path count alignment. Together, these strategies ensure SVGen generates high-quality, interpretable SVGs suitable for various applications.}
    \label{fig:double_column}
\end{figure*}
\section{Method}
In this study, we fine-tune large language models by integrating curriculum learning strategies, Chain-of-Thought reasoning, and reinforcement learning methods to achieve high-quality and interpretable SVG generation. The specific workflow is illustrated in Figure~\ref{fig:double_column}.

\subsection{Curriculum learning}
Curriculum learning\cite{bengio2009curriculum} is a training strategy that mimics the human learning process. The main idea is to begin with simple and easy-to-understand content, gradually transitioning to more complex and challenging tasks\cite{wang2021survey}. Based on this concept, our research implements the curriculum learning strategy through a phased training process. Initially, we train with structurally simple monochrome SVG data, enabling the model to grasp basic geometric shape recognition and path structure understanding. Building on this, we introduce more complex monochrome SVGs with nested structures to enhance the model's understanding of hierarchical relationships. As training progresses, we gradually incorporate complex SVG data with rich colors and textures, while correspondingly increasing the number of training iterations. This approach allows the model to effectively handle a greater diversity of colors and intricate details. Through this progressive training approach, the model not only significantly improves its understanding and generalization of complex patterns but also exhibits better performance in complex scenarios. More importantly, this strategy of training from simple to complex enhances the model's learning capability and accuracy, while effectively preventing overfitting and significantly boosting its adaptability in various application scenarios.

\subsection{Chain-of-Thought}
In recent years, Chain-of-Thought\cite{wei2022chain} has gained widespread attention as a strategy to enhance the reasoning ability of large language models. CoT breaks down problems into a series of iterative logical steps, enabling models to simulate human thought processes, which is especially effective in solving complex tasks such as mathematical reasoning and logical analysis. This study employs a CoT construction method based on knowledge distillation, utilizing a high-quality multimodal large model (GPT-4o) to analyze icons' content and generate structured descriptions.

We first send SVG icons and their textual descriptions into the multimodal model to automatically analyze the visual elements of the icons, including basic geometric shapes, spatial hierarchical relationships, color fill attributes, and special effects. During the structured description generation phase, the model outputs a standardized design process breakdown with 2 to 6 logical sequential steps, each containing clear information about shape, type, color, and relative position. For example: "1. Draw a red rectangle with a triangle on top to represent an open envelope" or "2. Add a smaller cream rectangle inside the envelope to represent a letter."

To ensure the professionalism and accuracy of the generated content, we design a rigorous prompt template and manually verify and correct all samples. The standardized descriptions are then paired with the original SVG code to create custom training samples, which include a complete reasoning chain and target code (for detailed information on CoT design, please refer to the supplementary materials). Experimental results show that this method significantly improves the interpretability of design logic and the consistency of code generation, particularly demonstrating strong handling capabilities for complex icons composed of multiple components. Through this CoT construction mechanism, the model not only understands the visual features of icons but also grasps the logical process from concept to implementation, laying an important foundation for high-quality SVG generation.

\subsection{Reinforcement learning}

Despite good performance, existing LLMs often produce incomplete SVGs and inaccurate path counts, which hinder visual quality and editability—especially for complex graphics.

To further optimize model performance regarding these issues, we design an integrity reward function and a path quantity matching reward function, and apply the Group Relative Policy Optimization \cite{shao2024deepseekmath} reinforcement learning method to guide the model towards generating SVG code with complete shapes and paths, as well as a path quantity that meets expectations.

\textbf{(1) Integrity Reward} \( R_{\text{int}}(S) \):
   We define a metric \( I(S) \) to assess SVG structural integrity: 1 if the generated SVG is successfully parsed and correctly closed by a standard SVG parser, 0 otherwise. The integrity reward function is expressed as:
   \[
   R_{\text{int}}(S) = \alpha I(S)
   \]
   where \( \alpha > 0 \) is a coefficient controlling the weight of the integrity reward. This mechanism encourages the model to maintain the comprehensiveness and consistency of SVG graphics under limited token conditions, reducing incomplete or fragmented graphics.

\textbf{(2) Path Number Matching Reward} \( R_{\text{match}}(S, S_{\text{gt}}) \): We design a mechanism to address the relationship between the generated path number and the reference path number, particularly focusing on ensuring that the complexity of the generated icons is equal to or greater than the reference icons, and rewarding such scenarios. The path number matching reward is defined as:

\[
R_{\text{match}}(S, S_{\text{gt}}) = \max\left(\beta, \beta \exp\left(-\gamma (N(S) - N(S_{\text{gt}}))\right)\right)
\]

Here, \( \beta > 0 \) regulates the overall reward magnitude, and \( \gamma > 0 \) modulates sensitivity to path number deviations. \( N(S) \) and \( N(S_{\text{gt}}) \) represent the path counts in the generated SVG and its corresponding reference SVG, respectively. With this design, when the generated path number is equal to or exceeds the reference count, the reward value automatically reaches \( \beta \); otherwise, it diminishes according to the discrepancy in path numbers. This approach encourages generating icons with a complexity at least equal to the reference icons while ensuring the generated result aligns closely with the ground truth in terms of path count. This reward mechanism ultimately enhances the structural realism and accuracy of the visual outputs, enabling the model to better balance complexity and precision.

\begin{table*}[htbp]
    \centering
    \renewcommand{\arraystretch}{1.2} % 可选：调整行距以适应需求
    \resizebox{0.99\textwidth}{!}{
        \begin{tabular}{cccccccccc}
            \toprule
            \multirow{2}{*}{\textbf{Model Type}} & \multirow{2}{*}{\textbf{Model}} & \multirow{2}{*}{\textbf{FID\textsubscript{$\downarrow$}}} & \multicolumn{2}{c}{\textbf{CLIPScore}} & \multirow{2}{*}{\textbf{HPS\textsubscript{$\uparrow$}}} & \multirow{2}{*}{\textbf{Aesthetic\textsubscript{$\uparrow$}}} & \multirow{2}{*}{\textbf{Avg. Token}} & \multirow{2}{*}{\textbf{Gen. Time\textsubscript{$\downarrow$}}} \\
            \cmidrule(lr){4-5}
             & & & \textbf{T2I\textsubscript{$\uparrow$}} & \textbf{I2I\textsubscript{$\uparrow$}} & & & & \\ 
             %[-2.5ex] % 可选：调整负偏移量使子标题更贴近
            \midrule
            \multirow{1}{*}{AR-Based} 
            & Iconshop\cite{wu2023iconshop} & 496.48 & 0.1932 & 0.6770 & 0.1756 & 4.6039 & 1356.61 & 13.48 s \\
            \midrule
            \multirow{3}{*}{Optimization Based} 
            & VectorFusion\cite{jain2023vectorfusion} & 145.26 & 0.2683 & 0.6734 & 0.1914 & 4.5182 & N/A & \(\approx\) 27.41 min  \\
            & DiffSketcher\cite{xing2023diffsketcher} & 235.25 & 0.2396 & 0.6759 & 0.1849 & 4.4711 & N/A & \(\approx\) 3.00 min  \\
            & SVGDreamer\cite{xing2024svgdreamer} & 124.34 & 0.2709 & 0.6778 & 0.1943 & 4.6991 & N/A & \(\approx\)  1.83 h  \\
            \midrule
            \multirow{7}{*}{LLM-Based} 
            & Qwen2.5 72B\cite{yang2024qwen2} & 44.63 & 0.2595 & 0.7364 & 0.1923 & 4.6838 & 267.21 & 11.85 s \\
            & Llama-3.3 70B\cite{grattafiori2024llama} & 51.96 & 0.2612 & 0.7323 & 0.1922 & 4.6623 & 213.08 & 9.25 s \\
            & GPT-4o-mini\cite{achiam2023gpt} & 51.43 & 0.2692 & 0.7490 & 0.1939 & 4.6669 & 258.20 & N/A \\
            & Gemini-2.5 Pro\cite{gemini_20} & 49.52 & 0.2684 & 0.7849 & \textbf{0.1964} & 4.7581 & 340.40 & N/A \\
            & Grok-3\cite{grok3_beta} & 47.59 & 0.2698 & 0.7705 & 0.1954 & 4.7245 & 232.43 & N/A \\
            &Deepseek V3\cite{liu2024deepseek}& 44.45 & 0.2677 & 0.7660 & 0.1951 & 4.7379 & 252.82 & N/A \\
            & GPT-4o\cite{achiam2023gpt}& 47.73 & \textbf{0.2737} & 0.7595 & 0.1948 & 4.6734 & 232.77 & N/A \\
            & Deepseek R1\cite{guo2025deepseek} & 44.54 & 0.2642 & 0.7746 & 0.1949 & 4.7341 & 268.31 & N/A \\
            \midrule
            \multirow{4}{*}{Ours}
            & Ours(Llama3.2-3B) & 32.54 & 0.2317 & 0.7935 & 0.1921 & 4.8305 & 1507.63 & 7.78 s \\
            & Ours(Qwen2.5-Inst-3B) & 30.67 & 0.2305 & 0.7953 & 0.1912 & 4.8543 & 1630.04 & \textbf{7.56 s} \\
            & Ours(Starcoder2-3B) & 30.52 & 0.2413 & \textbf{0.8125} & 0.1943 & 4.8838 & 1899.66 & 8.20 s \\
            & Ours(Qwen2.5-Coder-7B) & \textbf{27.82} & 0.2293 & 0.8080 & 0.1919 & \textbf{4.8924} & 1909.23 & 10.04 s \\
            \bottomrule
        \end{tabular}
    }
    \caption{Comparison of different models across various metrics. The evaluation content includes image quality, semantic information alignment effect, image style quality, SVG code length, and generation time, etc.}
    \label{tab:2}
\end{table*}

(3) The total reward function \( R(S, S_{\text{gt}}) \) combines the aforementioned rewards:
   \[
   R(S, S_{\text{gt}}) = R_{int} + R_{match}
   \]

By employing this reinforcement learning approach with GRPO, our model not only focuses on ensuring the integrity of the generation process but also effectively regulates the number of generated paths to closely match the structural characteristics of real data.

\section{Experiment}
To evaluate the performance of our proposed method for SVG generation tasks, we conduct comprehensive experiments on the constructed SVG-1M dataset. Totally 1,143 samples are randomly selected as an independent test set from the multicolored SVG, while the remaining data is used to train the model.

Our experimental framework employs a fine-tuning approach based on pretrained large language models, incorporating a progressive curriculum learning strategy for model optimization. The training process strictly follows a complexity-increasing learning trajectory: initial foundation training with 1 epoch each on monochrome SVG data (both simple and complex subsets), followed by advanced training with 3 epochs each on colored SVG data (simple and complex subsets), and concluding with additional fine-tuning using the 60,000 CoT-annotated samples for 3 epochs. 

For model configurations, we evaluate our framework on four basic architectures: StarCoder2-3B, Llama3.1-3B, Qwen2.5-3B-Instruct, and Qwen2.5-Coder-7B, performing full-parameter fine-tuning. 

All experiments are executed on 8×NVIDIA A800 GPUs using the AdamW optimizer (learning rate 4e-5), with a maximum sequence length of 8,000 tokens.
The coefficients of all rewards in the reinforcement learning phase are set to 1.

\subsection{Quantitative Analysis}
% Objectively and comprehensively evaluating the effectiveness and quality of text-to-SVG generation presents a significant challenge, as it involves cross-modal semantic information alignment. Therefore, we introduces multiple metrics from different sources in an attempt to conduct a comprehensive evaluation of the final generated outcomes from various perspectives. Specifically, for visual feature evaluation, the Fréchet Inception Distance (FID)\cite{heusel2017gans} is utilized as the primary metric. Additionally, a dual scoring mechanism based on the CLIP\cite{radford2021learning} model is introduced: CLIPScore-T2I is employed to evaluate the semantic consistency between text and SVG, while CLIPScore-I2I measures the visual similarity between the generated SVG and reference SVGs. Furthermore, an integrated aesthetic evaluation system is constructed by combining Human Preference Scores (HPS)\cite{wu2023human}with standardized aesthetic scores\cite{schuhmann2022}.

% In terms of efficiency assessment, a dual quantification standard is applied: one aspect measures the average time required for the model to generate SVGs, while the other calculates the average encoding length by employing a unified tokenizer (Qwen2.5) to standardize the processing of SVG code. This evaluation framework facilitates a systematic assessment encompassing generation quality, visual fidelity, and computational efficiency.
Evaluating text-to-SVG generation is challenging due to cross-modal semantics. We introduce multiple metrics for comprehensive evaluation. Fréchet Inception Distance (FID)\cite{heusel2017gans} is the primary visual metric. A CLIP-based\cite{radford2021learning} dual scoring mechanism includes CLIPScore-T2I for text-SVG semantic consistency and CLIPScore-I2I for visual similarity to reference SVGs. An integrated aesthetic evaluation system combines Human Preference Scores (HPS)\cite{wu2023human} with standardized aesthetic scores. For efficiency, we measure average generation time and average encoding length using a unified tokenizer (Qwen2.5). This framework systematically assesses generation quality, visual fidelity, and computational efficiency.

Table ~\ref{tab:2} presents a comprehensive comparison between our method and existing Text-to-SVG generation approaches, demonstrating superior performance across multiple key metrics. Our method achieves the highest Aesthetic Score, indicating exceptional visual quality in the generated SVGs. In terms of code complexity, our results significantly exceed those from LLM-based methods, reflecting enhanced detail representation and richer content generation. Furthermore, our approach shows substantially faster generation speeds compared to optimization-based models, highlighting its computational efficiency and practical applicability. These results collectively indicate that our method successfully balances visual quality, code complexity, and generation efficiency, establishing an effective solution for Text-to-SVG conversion tasks.

Further analysis reveals critical limitations in current approaches. Optimization-based methods tend to produce SVG code with unnecessarily high decimal precision and redundant path definitions, resulting in bloated file sizes and reduced editability. Meanwhile, LLM-based methods exhibit instability in generating icons with consistent dimensions and proportions when canvas size specifications are absent, significantly compromising their practical utility. In contrast, our method generates SVG code that maintains appropriate complexity while ensuring standardization and usability, delivering a more intuitive and efficient graphics generation experience for end-users.

% The comparative results clearly demonstrate our method's advantages in producing visually appealing, structurally sound, and efficiently generated SVG graphics from textual descriptions. By addressing the key limitations of existing approaches, our solution provides a robust framework for high-quality SVG generation that meets both aesthetic and functional requirements.
Comparative results clearly show our method excels at generating visually appealing, well-structured, and efficient SVG graphics from text. Our solution addresses existing approaches' limitations, offering a robust framework for high-quality SVG generation that satisfies both aesthetic and functional needs.

\subsection{Ablation study}
To systematically evaluate the synergistic effects of Curriculum Learning (CL), Chain-of-Thought (CoT), and Group Relative Policy Optimization (GRPO) reinforcement learning in the proposed SVGen, we design four progressively configured experimental groups: (1) Baseline model (without CL, with all training samples fed simultaneously); (2) CL-only; (3) CL+CoT joint training; and (4) Complete CL+CoT+GRPO framework. The experiments utilize the StarCoder2-3B and Qwen2.5-Instruct-3B models, employing a controlled variable methodology to rigorously isolate the individual contributions of each component.

% \begin{table}[h]
% \centering
% \resizebox{0.95\columnwidth}{!}{
% \begin{tabular}{lccc}
% \toprule
% Model & FID ($\downarrow$) & \multicolumn{2}{c}{CLIPScore} \\
% \cmidrule(lr){3-4}
%  & & T2I ($\uparrow$) & I2I ($\uparrow$) \\
% \midrule
% Ours(Starcoder2-3B) & \textbf{30.52} & \textbf{0.2413} & \textbf{0.8125} \\
% \midrule
% w/o Curriculum learning & 39.43 & 0.2351 & 0.7846 \\
% w/o Chain-of-Thought & 34.86 & 0.2385 & 0.8032 \\
% \midrule
% Ours(Qwen2.5-3B) & \textbf{30.67} & \textbf{0.2304} & \textbf{0.7953} \\
% \midrule
% w/o Curriculum learning & 53.46 & 0.2121 & 0.7603 \\
% w/o Chain-of-Thought & 42.39 & 0.2231 & 0.7869 \\
% \bottomrule
% \end{tabular}}
% \caption{Performance comparison of different models.}
% \end{table}
\begin{table}[h]
\centering
\resizebox{0.95\columnwidth}{!}{% 调整表格宽度为单栏宽度的95%
\footnotesize % 缩小字体大小
\begin{tabular}{lccccc} % 添加一列
\toprule
Model & FID \textsubscript{$\downarrow$} & HPS \textsubscript{$\uparrow$} & \multicolumn{2}{c}{CLIPScore} \\ % 调整列顺序
\cmidrule(lr){4-5}
 & & & T2I \textsubscript{$\uparrow$} & I2I \textsubscript{$\uparrow$} \\
\midrule
Ours(Starcoder2-3B) & \textbf{30.52} & \textbf{0.1943} & \textbf{0.2413} & \textbf{0.8125} \\ % 添加随机数据
\midrule
w/o Curriculum learning & 39.43 & 0.1925 & 0.2351 & 0.7846 \\ % 添加随机数据
w/o Chain-of-Thought & 34.86 & 0.1938 & 0.2385 & 0.8032 \\ % 添加随机数据
\midrule
Ours(Qwen2.5-Inst-3B) & \textbf{30.67} & \textbf{0.1912} & \textbf{0.2304} & \textbf{0.7953} \\ % 添加随机数据
\midrule
w/o Curriculum learning & 53.46 & 0.1835 & 0.2121 & 0.7603 \\ % 添加随机数据
w/o Chain-of-Thought & 42.39 & 0.1892 & 0.2231 & 0.7869 \\ % 添加随机数据
\bottomrule
\end{tabular}
}
\caption{Performance comparison of SVGen across different training procedures.}
\label{tab:3}
\end{table}

\textbf{Analysis of Curriculum Learning and Chain-of-Thought.} The experimental results presented in Table~\ref{tab:3} demonstrate that employing Curriculum Learning alone significantly enhances the model's comprehension of basic structures and shapes. For instance, when generating fundamental icons, models utilizing Curriculum Learning exhibit notable improvements in the FID metric compared to their non-curriculum counterparts, indicating enhanced capability in capturing and replicating the visual characteristics of real samples. However, when addressing the generation of complex colored icons, this approach still encounters challenges with inaccurate path quantity matching and suboptimal visual similarity.

% The introduction of the Chain-of-Thought mechanism has significantly enhanced the model's performance in  handling complex geometric compositions and precise color-shape matching. This improvement not only strengthens the structural integrity and visual consistency of the generated results but also effectively improves semantic accuracy in text-to-SVG generation. Specifically, the combined approach of curriculum learning with CoT demonstrates substantial score improvements across metrics including HPS and CLIPScore, comprehensively validating the method's effectiveness across three critical dimensions: enhancing generated image quality, ensuring precise alignment between images and their textual descriptions, and improving visual similarity between generated and reference images.

The introduction of the Chain-of-Thought mechanism significantly improves the model's ability to handle complex geometric compositions and precise color-shape matching. This enhancement boosts the structural integrity and visual consistency of generated results, while also improving semantic accuracy in text-to-SVG generation. Specifically, combining curriculum learning with CoT shows substantial score increases across metrics like HPS and CLIPScore. This broadly validates our method's effectiveness in three key areas: enhancing generated image quality, ensuring precise alignment between images and text, and improving visual similarity to reference images.

% \begin{figure}[htbp]
%     \centering
%     \includegraphics[width=0.45\textwidth]{photos/output.pdf}
%     \caption{dataset}
%     \label{fig:5}
% \end{figure}
\begin{figure}[!htbp]
    \centering
    \hspace*{-0.4cm} % 向左偏移 2cm
    \includegraphics[width=0.45\textwidth]{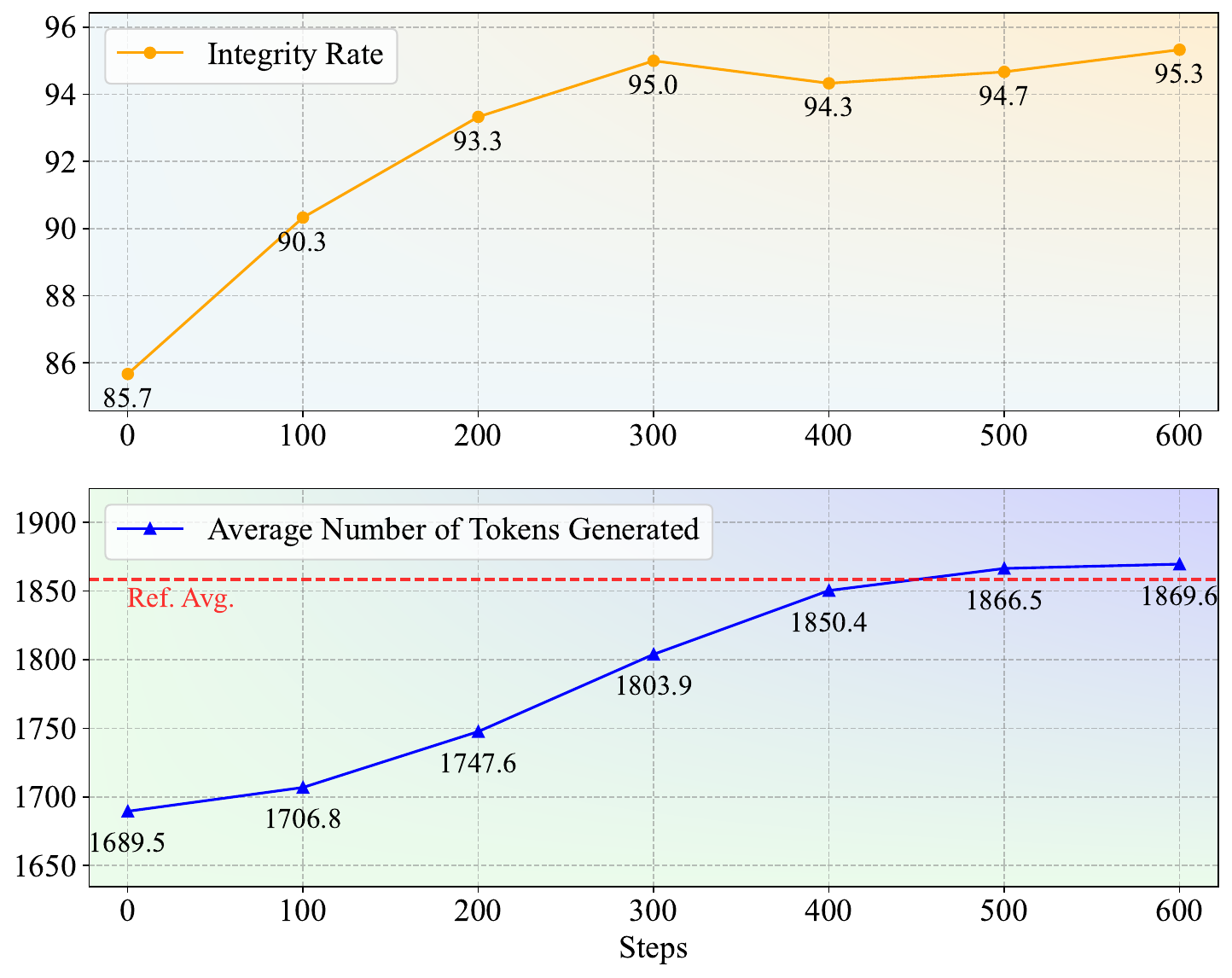}
    \caption{The structural integrity and path quantity changes during the reinforcement learning process of StarCoder2-3B (tested every 100 steps).}
    \label{fig:5}
\end{figure}

\begin{figure*}[htbp]
    \centering
    \includegraphics[width=\textwidth]{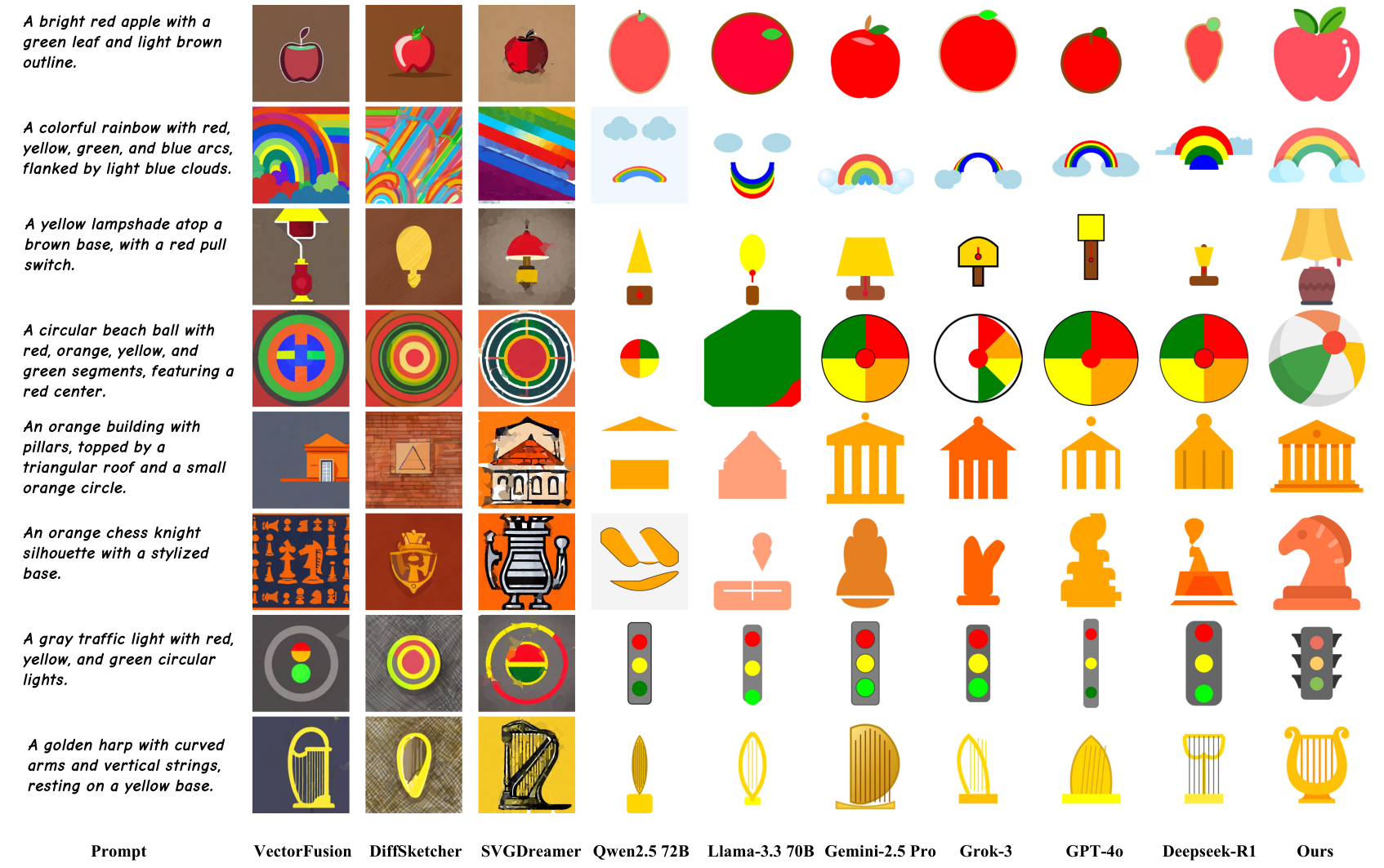}
    \caption{Qualitative comparison between the proposed SVGen and both optimization-based and LLM-based methods}
    \label{fig:fig4}
\end{figure*}

\textbf{Analysis of Reinforcement Learning.} During the experiments, we specifically select approximately 1,600 cases from the supervised fine-tuning phase that either fail to generate complete structures or produce results with significantly lower complexity than reference icons. These cases are exclusively used for reinforcement learning training based on the GRPO algorithm. By targeting these challenging samples, we aim to enhance the model's performance in complex scenarios, particularly in improving structural completeness and path quantity matching capabilities. The results are presented in Figure ~\ref{fig:5}. The reinforcement training consists of 600 total steps. To evaluate model progression, we save checkpoints every 100 steps and conduct evaluations using a predefined test set of 300 samples. The evaluation focuses on two key metrics: completion rate of generated results and average number of generated paths. Experimental data reveal consistent improvements in both metrics throughout the training process. The completion rate shows steady growth while the average path count progressively approaches and ultimately matches the reference icon's path complexity. Applying the GRPO algorithm with a custom reward yields two key benefits: improved structural integrity and increased icon complexity. The findings confirm the effectiveness of reinforcement learning in addressing SVG generation challenges.

% \begin{figure*}[htbp]
%     \centering
%     \includegraphics[width=\textwidth]{photos/10.pdf}
%     \caption{Qualitative comparison between the proposed SVGen and both optimization-based and LLM-based methods}
%     \label{fig:fig4}
% \end{figure*}
\begin{table}[htbp]
\centering
\resizebox{\columnwidth}{!}{% 将表格宽度调整为单栏宽度
\begin{tabular}{l|c|c|c|c}
\toprule
\textbf{Metric/Method} & \textbf{GPT-4o} & \textbf{DeepSeek R1} & \textbf{Gemini-2.5 Pro} & \textbf{SVGen} \\
\midrule
Semantic Match & 5.54 & 6.12 & 6.94 & 9.40  \\
Visual Quality &  5.68 & 6.20 & 6.24 & 9.88 \\
Usability & 5.62  & 6.16 & 6.58& 9.64  \\
\bottomrule
\end{tabular}
}
\caption{Results of Human Evaluation. Assign scores to each contrast sample based on their ranking, with a maximum score of ten. Finally, calculate the average score.}
\label{tab:tab4}
\end{table}
\subsection{Qualitative Analysis}

Figure ~\ref{fig:fig4} presents a comparison of the generation results of our SVGen method with those of optimization-based approaches and LLM-based methods. As can be seen from the figure, our method outperforms other LLM-based and optimization-based approaches in terms of both accuracy and aesthetic appeal in SVG generation. The images produced by our method are more aligned with the descriptions, feature proper handling of details, and exhibit superior overall quality.

% \begin{table}[htbp]
% \centering
% \resizebox{\columnwidth}{!}{% 将表格宽度调整为单栏宽度
% \begin{tabular}{l|c|c|c|c}
% \toprule
% \textbf{Metric/Method} & \textbf{GPT-4o} & \textbf{DeepSeek R1} & \textbf{Gemini-2.5 Pro} & \textbf{SVGen} \\
% \midrule
% Semantic Match & 5.54 & 6.12 & 6.94 & 9.40  \\
% Visual Quality &  5.68 & 6.20 & 6.24 & 9.88 \\
% Usability & 5.62  & 6.16 & 6.58& 9.64  \\
% \bottomrule
% \end{tabular}
% }
% \caption{Results of Human Evaluation. Assign scores to each contrast sample based on their ranking, with a maximum score of ten. Finally, calculate the average score.}
% \label{tab:tab4}
% \end{table}

\textbf{Human Evaluation Experiment Design:} Although metrics like CLIPScore and HPS offer objective evaluation, they overlook key aspects such as semantic consistency, visual appeal, and practical value. To address this, we conducted a blind review with 15 UI designers and 15 front-end engineers, focusing on three core aspects: text-icon semantic match, aesthetic quality, and practical usability. As shown in Table 4, SVGen significantly outperformed other models. This comprehensive evaluation ensures both technical soundness and professional design quality. For the complete evaluation scheme design, implementation details, and extended analysis results, please refer to the supplementary materials.
% Despite the fact that quantitative metrics such as CLIPScore and HPS can provide objective performance measurements, they often fail to comprehensively reflect the critical dimensions of the generated outcomes in terms of semantic consistency, aesthetic quality, and practical application value. To address this limitation, we have specially introduced a blind evaluation process involving 15 senior UI designers and 15 experienced front-end engineers, focusing on following three core aspects: (1)The semantic matching degree between text and icons; (2)The visual fidelity and aesthetic appeal;(3)The practical value within professional workflows.

% This study systematically evaluated four models (SVGen, GPT-4o, DeepSeek-R1, and Gemini2.5-Pro). The results are shown in Table ~\ref{tab:tab4}, where our SVG generation method significantly outperforms other large language model-based approaches in terms of performance. This multidimensional evaluation approach not only validates the technical performance of the models but also ensures that the generated outcomes meet professional design standards, thereby guiding model optimization to enhance generation quality and user satisfaction. For the complete evaluation scheme design, implementation details, and extended analysis results, please refer to the supplementary materials.

\section{Conclusions}
In this study, by integrating curriculum learning, Chain of Thought, and Reinforcement Learning Policy, we significantly enhanced the performance of large language models in generating high-quality SVG icons. The experimental results indicate that the curriculum learning strategy aids the model in swiftly acquiring fundamental structure recognition abilities, CoT improves the model's comprehension and generation quality for complex tasks, and GRPO further refines the integrity and path count matching of the generated outcomes.

% Looking ahead, we will continue to explore how to expand this framework to accommodate more diverse application scenarios. On one hand, we plan to introduce more types of training data, encompassing icons of varying styles and complexities, to enhance the model's generalization capability. On the other hand, we are considering applying this approach to other forms of vector graphic generation tasks, such as CAD design or map drawing, to explore its potential in a broader range of applications. Through these efforts, we aim to develop more intelligent and adaptable generative systems that cater to the growing demands of design and propel technological advancements in related fields.
Looking ahead, we will continue to expand this framework to support more diverse applications. On one hand, we plan to introduce training data with varying icon styles and complexities to enhance generalization. On the other hand, we aim to explore its use in other vector graphic tasks such as CAD design and map drawing. Through these efforts, we seek to build more intelligent and adaptable generative systems that meet growing design needs and advance related technologies.

% We also look forward to collaborating with other research teams to jointly address existing challenges in the technology, such as enhancing generation efficiency and handling larger datasets. The ultimate goal is to build a universal and efficient generative framework that not only excels in SVG icon generation but is also applicable to a wide array of creative and industrial design tasks.

\begin{acks}
This work was supported in part by the Natural Science Foundation of China under Grant 62306241, and in part by grants from the Innovation Foundation for Doctor Dissertation of Northwestern Polytechnical University (No.CX2025109).
\end{acks}

%%
%% The acknowledgments section is defined using the "acks" environment
%% (and NOT an unnumbered section). This ensures the proper
%% identification of the section in the article metadata, and the
%% consistent spelling of the heading.
% \begin{acks}
% To Robert, for the bagels and explaining CMYK and color spaces.
% \end{acks}

%%
%% The next two lines define the bibliography style to be used, and
%% the bibliography file.
\bibliographystyle{ACM-Reference-Format}
\bibliography{ref}
%
% If your work has an appendix, this is the place to put it.

\end{document}

\title{SVGen: Interpretable Vector Graphics Generation with Large Language Models}

\renewcommand\footnotetextcopyrightpermission[1]{}
\settopmatter{printacmref=false} %remove ACM reference format

\begin{teaserfigure}
    \centering
    \includegraphics[width=\textwidth]{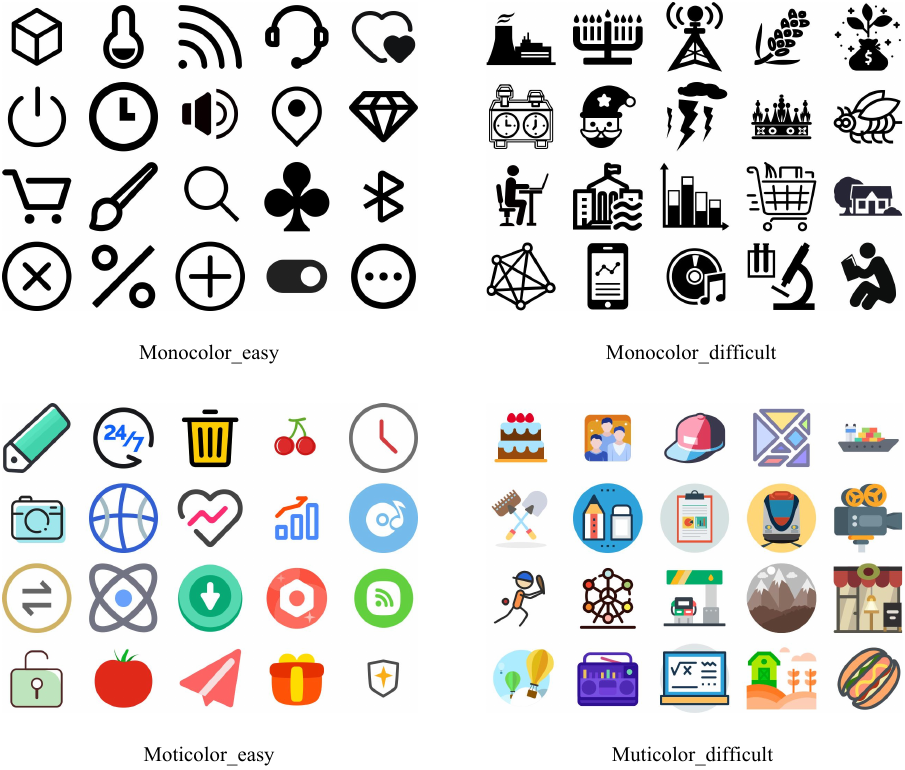}
    \caption{Data samples under different difficulty categories from SVG-1M}
    \label{fig:sample_diff_level}
\end{teaserfigure}
\maketitle

\section{Data processing pipeline}
\subsection{Data Collection}

    Data Collection. We collected SVG icons from Iconfont, a professional vector icon repository offering a diverse range of resources, including emojis, illustrations, fonts, and office icons. All icons on this platform strictly follow unified specifications: they use standard SVG namespaces, adhere to a 1024×1024 pixel canvas size, and maintain clean path structures (free of redundant metadata). These standardized features make Iconfont an ideal benchmark dataset for research. Our dataset comprises approximately 500,000 SVG icons, including 400,000 monochrome and 100,000 colored icons. We quantify computational complexity based on the number of path commands, then further classify the icons into four categories according to color types: Monochrome Simple (single-color icons with fewer commands and clean structures), Monochrome Complex (single-color icons with more commands or intricate details), Multicolor Simple (icons with clearly separated color regions and fewer commands), and Multicolor Complex (icons featuring gradients, blends, or multi-layer structures).

\subsection{Data standardization.}
In the representation of SVG graphics, various elements are typically defined using different basic tags, such as those for rectangles, circles, and custom paths. This diverse tagging system leads to inconsistencies in the representation methods and increases the complexity of subsequent processing and operations.

\begin{figure}[!htbp]
    \centering
    \includegraphics[width=0.48\textwidth]{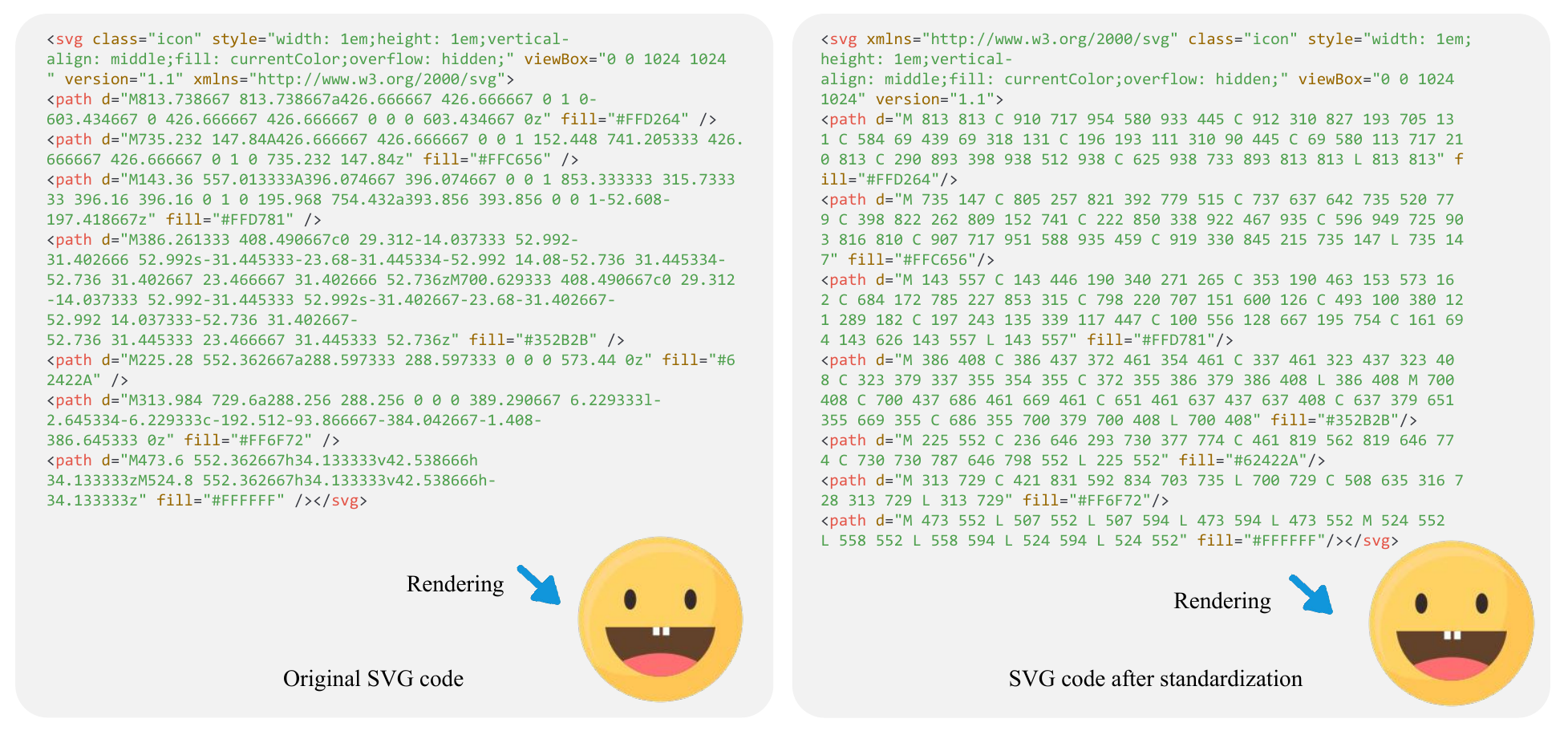}
    \caption{A comparison of the code and visual quality before and after processing.}
    \label{fig:6}
\end{figure}

To address this issue, we adopt a path unification approach, converting all graphic elements into standard elements. The converted paths use a minimal set of drawing instructions, including M (move to), L (line to), and C (cubic Bézier curve), as shown in Table ~\ref{tab:svg_commands}. Through this transformation, all graphic elements—including basic geometric shapes—can be accurately represented using a combination of Bézier curves and straight line segments. Ultimately, the SVG document is structured to contain only namespace declarations and collections of paths. Figure ~\ref{fig:6} illustrates examples of the comparison between the code and visual quality before and after processing.

\begin{table}[!htbp]
    \centering
    \renewcommand{\arraystretch}{1.5} % 增加行高
    \setlength{\tabcolsep}{4pt} % 调整列间距
    \begin{tabularx}{\columnwidth}{>{\centering\arraybackslash}m{1.8cm} >{\centering\arraybackslash}m{1.2cm} >{\centering\arraybackslash}m{2cm} >{\centering\arraybackslash}m{2cm}}
    \hline
    Name & Symbol & Arguments & Visualization \\ 
    \hline
    Move To & M & $(x_1, y_1), (x_2, y_2)$ & \includegraphics[height=1cm,width=1.5cm,keepaspectratio]{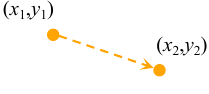} \\
    \hline
    Line To & L & $(x_1, y_1), (x_2, y_2)$ & \includegraphics[height=1cm,width=1.5cm,keepaspectratio]{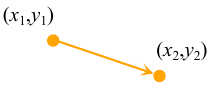} \\
    \hline
    Cubic Bézier & C & \begin{tabular}{@{}c@{}}$(x_1, y_1), (q_1^x, q_1^y),$ \\ $(q_2^x, q_2^y), (x_2, y_2)$\end{tabular} & \includegraphics[height=1cm,width=1.5cm,keepaspectratio]{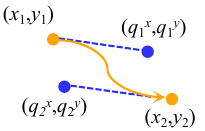} \\
    \hline
    \end{tabularx}
    \caption{List of simplified SVG commands with their names, symbols, arguments, and visualizations.}
    \label{tab:svg_commands}
\end{table}

% \begin{figure}[htbp]
%     \centering
%     \includegraphics[width=0.45\textwidth]{img/zw.png}
%     \caption{xxxxx}
%     \label{fig:svg_code}
% \end{figure}

\section{Data Inference}

All icons undergo rasterization at 200×200 pixel resolution, with GPT-4o subsequently generating comprehensive textual descriptions that capture both geometric features and semantic meaning with sufficient detail for accurate visual reconstruction. The prompt used for generate comprehensive textual descriptions is shown in Figure~\ref{fig:desc_gen_pro_4o}.

\begin{figure}[htbp]
    \centering
    \begin{tcolorbox}[
        colframe=blue!30!black,
        colback=blue!5!white,
        boxrule=1pt,
        arc=4pt,
        outer arc=4pt,
        boxsep=5pt,
        left=6pt,
        right=6pt,
        top=6pt,
        bottom=6pt,
        fonttitle=\bfseries,
        title=Prompt: GPT-4o image description,
        ]
    \small
    You are an image analysis assistant. Please help me briefly describe the provided icon. Make sure that everyone can clearly imagine the content and style of this icon through your description. The description text is as short and clear as possible, and no more than 30 Tokens is allowed.
    \textit{Goal: Create robust SVG representations for research purposes.}
    \end{tcolorbox}
    \caption{Prompts for generating comprehensive SVG icon descriptions using GPT-4o}
    \label{fig:desc_gen_pro_4o}
\end{figure}

For monochromatic icons, we supplement this with lightweight visual language models like Qwen2.5-VL-7B to produce basic structural descriptions, and the prompt used for basic descriptions is shown in Figure~\ref{fig:desv_gen_prompt_qwen}.

\begin{figure}[htbp]
    \centering
    \begin{tcolorbox}[
        colframe=blue!30!black,
        colback=blue!5!white,
        boxrule=1pt,
        arc=4pt,
        outer arc=4pt,
        boxsep=5pt,
        left=6pt,
        right=6pt,
        top=6pt,
        bottom=6pt,
        fonttitle=\bfseries,
        title=Prompt: Qwen2.5-VL-7B image description,]
    \small
    You are an image analysis assistant. Use the most relevant word or phrase to describe this icon.
    \textit{Goal: Create robust SVG representations for research purposes.}
    \end{tcolorbox}
    \caption{Prompts for generating basic descriptions of the SVG icons using Qwen2.5-VL-7B}
    \label{fig:desv_gen_prompt_qwen}
\end{figure}

% The color icon dataset undergoes significant expansion through algorithmic transformations including path recombination and adaptive color palette substitution, systematically increasing the original 100,000 samples to 203,205 distinct variations while preserving core visual characteristics. This approach ensures our dataset captures both high-fidelity visual information and robust structural variations essential for comprehensive model training.

\begin{figure*}[t]
    \centering
    \begin{tcolorbox}[title=Prompt and CoT, coltitle=black, colbacktitle=gray!30!white, colframe=black, width=\textwidth, boxrule=1mm]
    \begin{tcolorbox}[colback=gray!10!white]
    \begin{minipage}{0.82\textwidth}  % Adjust width to allow more space for the text
    \textbf{Prompt:}\\
    Please generate an SVG icon that meets the following description: A yellow smiling face with red cheeks, blue tears, and a pink tongue.
    \vspace{6pt}
    
    \textbf{CoT:}\\
    1. Create a large yellow circle for the face and add a slightly darker yellow crescent on the left side for shading. 
    
    2. Draw two small, curved maroon lines for the eyes and position them symmetrically near the top of the circle. 
    
    3. Add two blue teardrop shapes, one on each side of the face, angled downwards to represent tears of joy. 
    
    4. Design a wide, open mouth using a maroon semicircle and place it in the lower center of the face. 
    
    5. Insert a smaller, red semicircle inside the mouth to depict the tongue, and add a white rectangle above it to represent teeth.
    \end{minipage}
    \hfill
    \begin{minipage}{0.15\textwidth}  % Adjust width for the image
    \includegraphics[width=\linewidth]{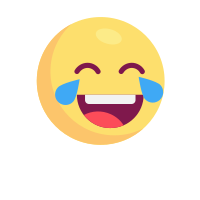} % Insert your image path here
    \end{minipage}
    \end{tcolorbox}
    
    % Second Question and Answer Box
    \begin{tcolorbox}[colback=gray!10!white]
    \begin{minipage}{0.82\textwidth}  % Adjust width to allow more space for the text
    \textbf{Prompt:}\\
    Please generate an SVG icon that meets the following description: Red bird on a brown branch with a yellow body and a green eye.
    \vspace{6pt}
    
    \textbf{CoT:}\\
    1. Create a red circle for the bird's head and a smaller red circle for the bird's eye, filling the eye with a brown color. 
    
    2. Draw a red oval for the bird's body, ensuring it overlaps with the head. 
    
    3. Add a yellow circle behind the bird's body to represent the sun. 
    
    4. Construct two yellow rectangles for the bird's legs and place them below the body. 
    
    5. Design a brown horizontal line for the branch and add two small brown rectangles at the ends to represent the branch's supports.
    \end{minipage}
    \hfill
    \begin{minipage}{0.15\textwidth}  % Adjust width for the image
    \includegraphics[width=\linewidth]{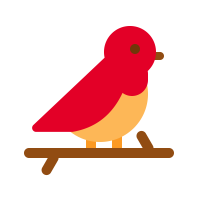} % Insert your image path here
    \end{minipage}
    \end{tcolorbox}
    
    % Third Question and Answer Box
    \begin{tcolorbox}[colback=gray!10!white]
    \begin{minipage}{0.82\textwidth}  % Adjust width to allow more space for the text
    \textbf{Prompt:}\\
    Please generate an SVG icon that meets the following description: A pink sunburst with a yellow circle, featuring a brown-faced smiley wearing black sunglasses.
    \vspace{6pt}
    
    \textbf{CoT:}\\
    1. Create a large circle with a pink fill to form the outer gear-like shape, adding wavy edges to mimic the gear's teeth. 
    
    2. Draw a smaller circle with a yellow fill and place it centrally within the pink gear shape to represent the sun. 
    
    3. Add two rounded rectangles with a brown fill for the sunglasses, positioning them symmetrically on the yellow circle. 
    
    4. Connect the sunglasses with a small horizontal rectangle in brown to form the bridge of the glasses. 
    
    5. Draw a small curved line with a brown fill below the sunglasses to create a smiling mouth.
    \end{minipage}
    \hfill
    \begin{minipage}{0.15\textwidth}  % Adjust width for the image
    \includegraphics[width=\linewidth]{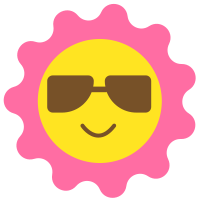} % Insert your image path here
    \end{minipage}
    \end{tcolorbox}
    \end{tcolorbox}
    \caption{Samples of chain-of-thought data pairs from SVG-1M.}
    \label{fig:cot_examples}
\end{figure*}

As for the CoT data, we use the prompt provided in Figure~\ref{fig:cot_gen_prompt_4o} to obtain results from GPT-4o.
Figure~\ref{fig:cot_examples} provides example paired data of CoT for reference.

\begin{figure}[htbp]
    \centering
    \begin{tcolorbox}[
        colframe=blue!30!black,
        colback=blue!5!white,
        boxrule=1pt,
        arc=4pt,
        outer arc=4pt,
        boxsep=5pt,
        left=6pt,
        right=6pt,
        top=6pt,
        bottom=6pt,,
        fonttitle=\bfseries,
        title=Prompt: GPT-4o Chain-of-Thought reasoning,]
    \small
    You are an image analysis assistant. I need to design an SVG based on this icon and analyze its components. Describe the design process with numbered points like '1...2...', connecting components into a complete process. Ensure to use numbers for points and limit to 2-6. Fewer points are better if the process is clear. Include shape names and fill colors. Make sure each sentence is complete.
    \textit{Goal: Create robust SVG representations for research purposes.}
    \end{tcolorbox}
    \caption{Prompts for constructing detailed CoT steps for SVG icon drawing}
    \label{fig:cot_gen_prompt_4o}
\end{figure}

\section{Human Evaluation}
This study designed a systematic human evaluation scheme, implementing a professional online scoring system through Streamlit to standardize and automate the evaluation process. We invited 15 senior UI designers and 15 experienced front-end engineers to form an evaluation panel. They rigorously assessed 200 sets of SVG icons generated by four models—SVGen, GPT-4o, DeepSeek-R1, and Gemini2.5-Pro—from 50 representative text prompts, covering common icon types such as tools, social media, indicators, and various complexity levels.

The evaluation system, built on the Streamlit platform, enabled automatic task allocation, standardized scoring interfaces, and real-time data collection and verification. The assessment followed a strict double-blind testing procedure, with each evaluator independently completing professional judgments on the following three dimensions:
\begin{itemize}
    \item \textbf{Semantic Match:} Assessing whether the generated icons accurately convey the design intent, including accuracy in element composition, metaphor expression, and functional indication;
    \item \textbf{Visual Quality:} Evaluating smoothness of lines, color coordination, proportion balance, and overall aesthetic appeal from a professional perspective;
    \item \textbf{Practical Value:} Considering the usability of icons in real projects, including technical implementation difficulty, style compatibility, and adherence to design standards.
\end{itemize}

\begin{figure}[htbp]
    \centering
    \includegraphics[width=0.5\textwidth]{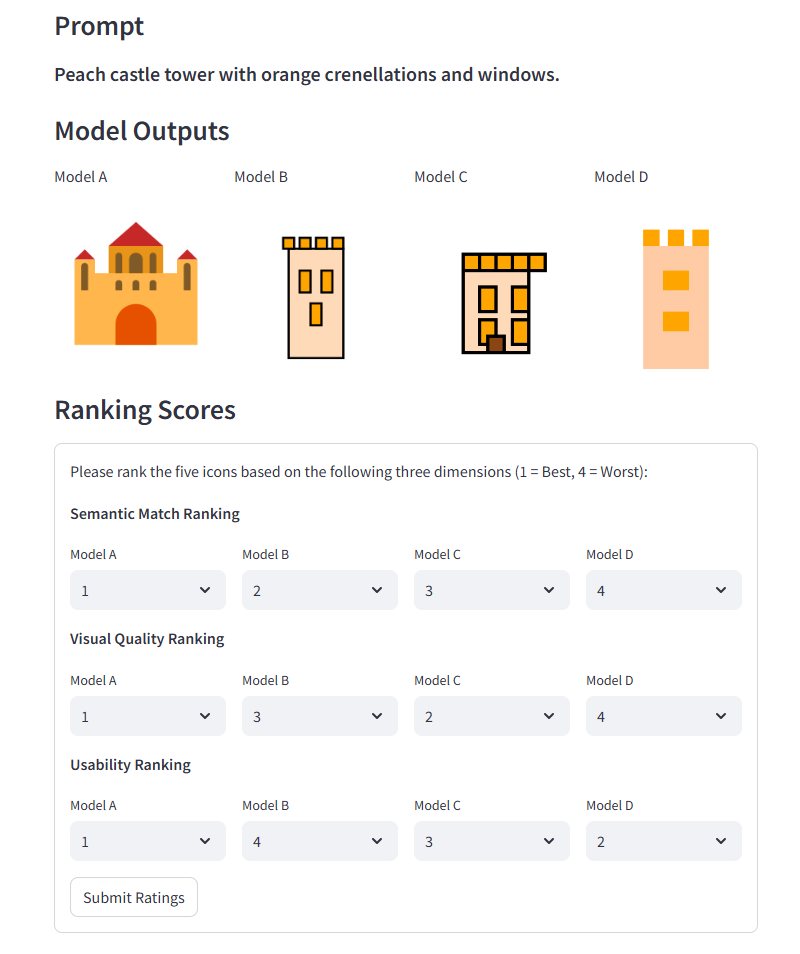}
    \caption{Human evaluation interface}
    \label{fig:human_eval_ui}
\end{figure}

Evaluators ranked the icons based on these three criteria, and the system used a weighted ranking score method (10/8/6/4 points) to automatically calculate the comprehensive scores for each model. The complete evaluation interface design is shown in Figure~\ref{fig:human_eval_ui}.

\section{Results of SVGen}
\begin{figure}[htbp]
    \centering
    \includegraphics[width=0.45\textwidth]{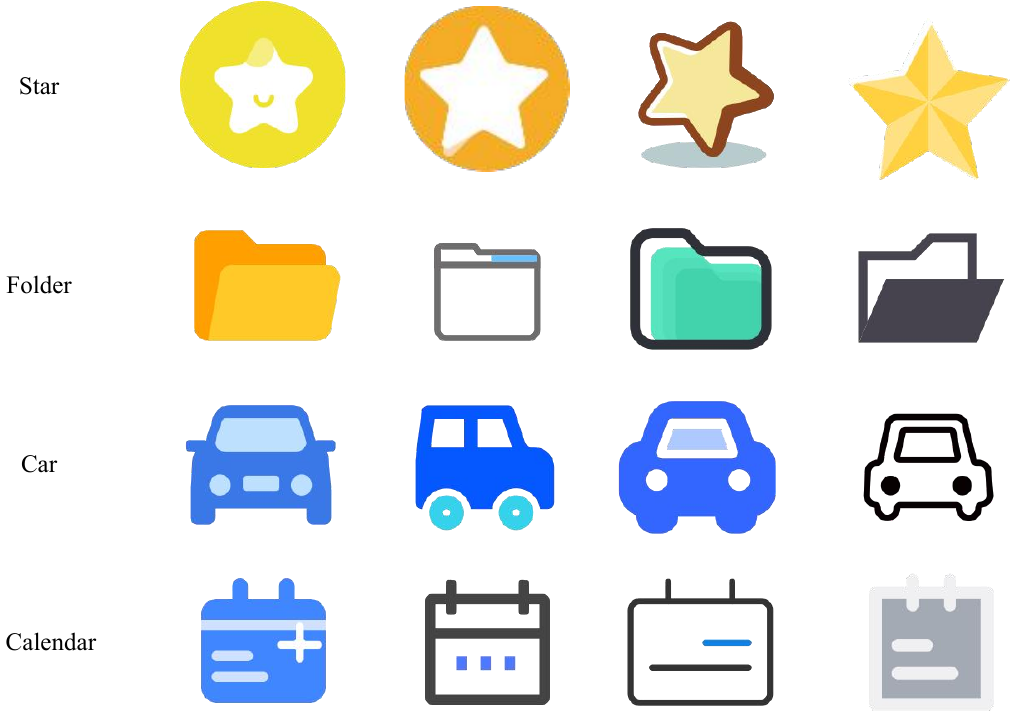}
    \caption{Testing the diversity of generated results.}
    \label{fig:res_diversity}
\end{figure}

\begin{figure*}[!htbp]
    \centering
    \includegraphics[width=0.9\textwidth]{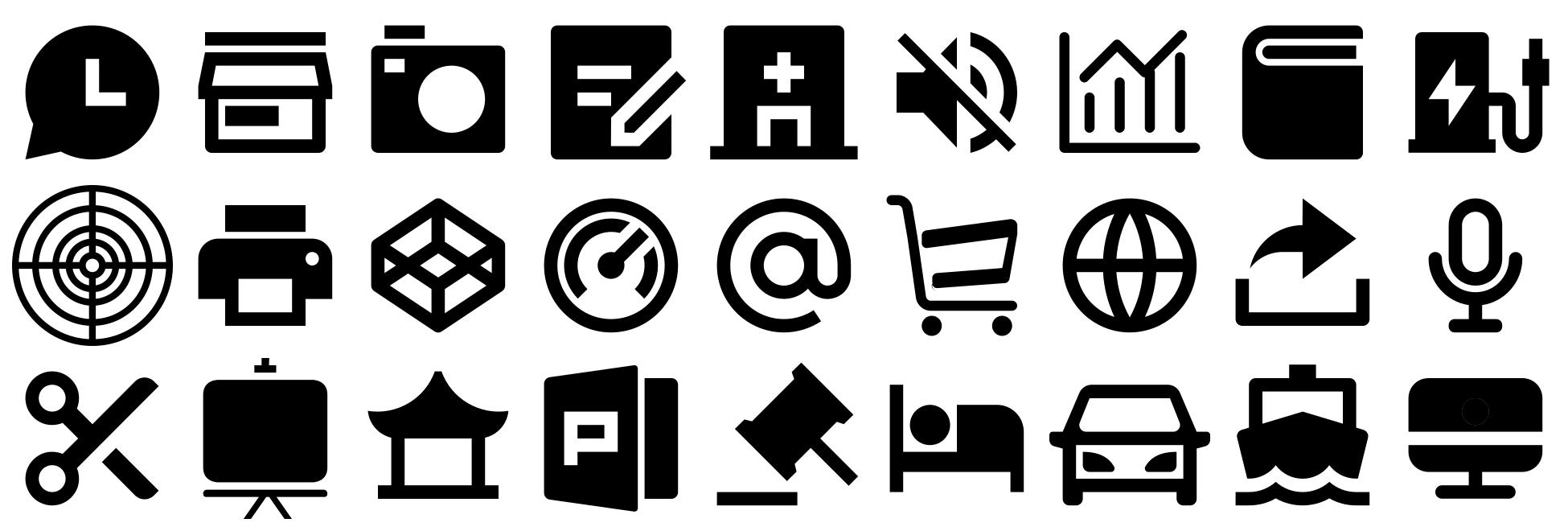}
    \caption{Monochrome results generated by SVGen}
    \label{fig:svgen_res_mon}
\end{figure*}

\begin{figure*}[htbp]
    \centering
    \includegraphics[width=0.9\textwidth]{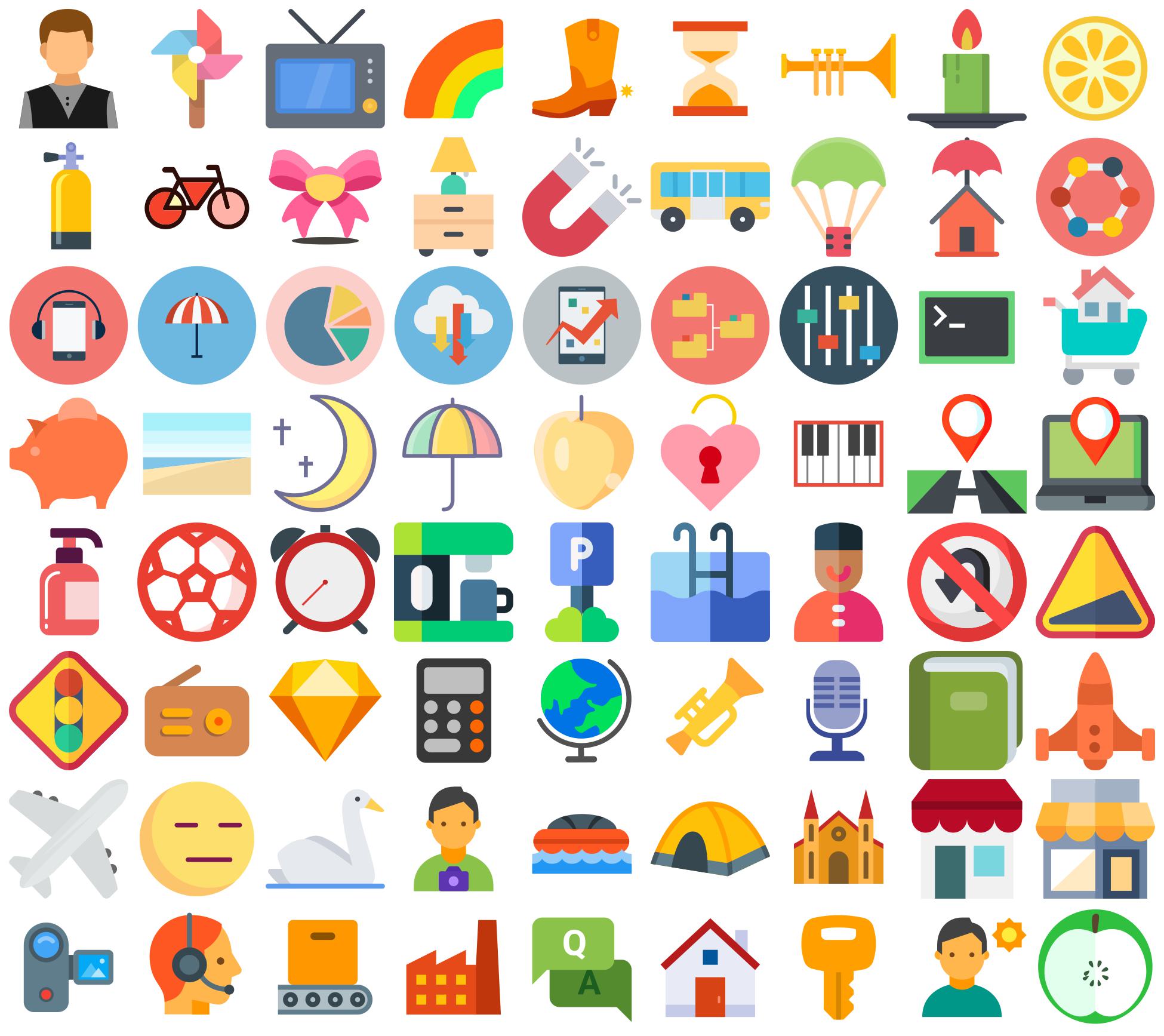}
    \caption{Multicolored results generated by SVGen}
    \label{fig:svgen_res_multi}
\end{figure*}

Figure~\ref{fig:res_diversity} shows the diversity of generated results of SVGen. 
Each row represents different results generated multiple times using the same target text as input.
Figure~\ref{fig:svgen_res_mon} and Figure~\ref{fig:svgen_res_multi} show the generated results of Monochrome and Multicolored SVGs.

\clearpage